\documentclass[runningheads]{llncs}

\usepackage{amsmath}
\usepackage{multirow}
\usepackage{subcaption}
\usepackage{tikz}
\usetikzlibrary{positioning, arrows, shapes,snakes, calc}
\usepackage{ifthen}
\usepackage{pgfplots, pgfplotstable}
\usepackage{graphicx}
\usepackage{booktabs}
\usepackage{wrapfig}
\usepackage{amssymb}
\usepackage{url}

\usepackage{makecell}
\usepackage[T1]{fontenc}
\usepackage{graphicx}

\usepackage{changepage} 
\usepackage[export]{adjustbox} 
\usepackage[misc]{ifsym}

\newcommand{\dset}[0]{\mathcal{D}}                          
\newcommand{\dsettrain}[0]{\dset_{\text{train}}}            
\newcommand{\dsetval}[0]{\dset_{\text{val}}}                
\newcommand{\dsettest}[0]{\dset_{\text{test}}}              

\newcommand{\loss}[0]{\mathcal{L}}              
\newcommand{\weights}[0]{\boldsymbol{\theta}}                            

\newcommand{\conf}[0]{\pmb{\lambda}}            
\newcommand{\hist}[0]{\dset_{\text{hist}}}                  

\newcommand{\seqidx}{\mathit{i}}

\newcommand{\target}{\mathbf{y}}
\newcommand{\prediction}{\hat{\mathbf{y}}}
\newcommand{\feature}{\mathbf{x}}
\newcommand{\featurespast}{\mathbf{x}^{(p)}}
\newcommand{\featuresfuture}{\mathbf{x}^{(f)}}
\newcommand{\seqlength}{T}

\newcommand{\dsetseq}{\dset_{\seqidx}}

\newcommand{\targetseq}[2][\seqidx]{\target_{#1, #2}}

\newcommand{\featurespastseq}[2][\seqidx]{\featurespast_{#1, #2}}
\newcommand{\featuresfutureseq}[2][\seqidx]{\featuresfuture_{#1, #2}}

\newcommand{\horizon}{H}

\newcommand{\seqlengthseq}[1][\seqidx]{\seqlength_{#1}}
\newcommand{\numseq}{N}

\newcommand{\observedintervalseq}{1:\seqlengthseq}
\newcommand{\forecastintervalseq}{\seqlengthseq +1 : \seqlengthseq + \horizon}

\newcommand{\observedinterval}{1:\seqlength}
\newcommand{\forecastinterval}{\seqlength +1 : \seqlength + \horizon}

\newcommand{\traininterval}{1:\seqlength - \horizon}
\newcommand{\valinterval}{\seqlength - \horizon +1: \seqlength}
\newcommand{\testinterval}{\seqlength + 1: \seqlength+ \horizon}
\newcommand{\allinterval}{1: \seqlength + \horizon}

\newcommand{\hiddenstate}{\mathbf{hx}}

\begin{document}

\title{Efficient Automated Deep Learning for\newline \  Time Series Forecasting}




%

\author{Difan Deng\Letter\inst{1} \and Florian Karl\inst{2,3} \and \newline Frank Hutter\inst{4,5} \and Bernd Bischl\inst{2,3} \and Marius Lindauer\inst{1}}
\authorrunning{D. Deng et al.}

\institute{Leibniz University Hannover, Hannover, Germany \\ \email{deng@tnt.uni-hannover.de} \and
Ludwig-Maximilian University, Munich, Germany  \and
Fraunhofer Institute for Integrated Circuits (IIS), Erlangen, Germany \and
University of Freiburg, Freiburg, Germany \and
Bosch Center for Artificial Intelligence, Renningen, Germany  
}

\newcommand{\system}[0]{\texttt{Auto-PyTorch-TS}}
\toctitle{Efficient Automated Deep Learning for Time Series Forecasting}
\tocauthor{Difan~Deng} 
\tocauthor{Florian~Karl}
\tocauthor{Frank~Hutter}
\tocauthor{Bernd~Bischl}
\tocauthor{Marius~Lindauer}

\maketitle              
%
\begin{abstract}
  Recent years have witnessed tremendously improved efficiency of Automated Machine Learning (AutoML), especially Automated Deep Learning (AutoDL) systems, but recent work focuses on tabular, image, or NLP tasks. So far, little attention has been paid to general AutoDL frameworks for time series forecasting, despite the enormous success in applying different novel architectures to such tasks. In this paper, we propose an efficient approach for the joint optimization of neural architecture and hyperparameters of the entire data processing pipeline for time series forecasting. In contrast to common NAS search spaces, we designed a novel neural architecture search space covering various state-of-the-art architectures, allowing for an efficient macro-search over different DL approaches. To efficiently search in such a large configuration space, we use Bayesian optimization with multi-fidelity optimization. We empirically study several different budget types enabling efficient multi-fidelity optimization on different forecasting datasets. Furthermore, we compared our resulting system, dubbed \system, against several established baselines and show that it significantly outperforms all of them across several datasets. 
\end{abstract}



\section{Introduction}
Time series (TS) forecasting plays a key role in many business and industrial problems, because an accurate forecasting model is a crucial part of a data-driven decision-making system. 
Previous forecasting approaches mainly consider each individual time series as one task and create a local model~\cite{assimakopoulos2000theta,box2015timeARIMA,hyndman2008forecastingETS}. In recent years, with growing dataset size and the ascent of Deep Learning (DL), research interests have shifted to global forecasting models that are able to learn information across all time series in a dataset collected from similar sources~\cite{godahewa2021monash,MAKRIDAKIS202054M4}. Given the strong ability of DL models to learn complex feature representations from a large amount of data, there is a growing trend of applying new DL models to forecasting tasks~\cite{lim2019tft,OreshkinICLR2020NBEATS,Flunkert17DeepAR,wen2017multi}.

Automated machine learning (AutoML) addresses the need of choosing the architecture and its hyperparameters depending on the task at hand to achieve peak predictive performance. The former is formalized as neural architecture search (NAS)~\cite{Elsken19NAS} and the latter as hyperparameter optimization (HPO)~\cite{feurer2019Hyperparameter}. Several techniques from the fields of NAS and HPO have been successfully applied to tabular and image benchmarks~\cite{Erickson20AutoGluon,FeurerNIPS15ASKL,jin2019autokeras,zimmertpami21aAPT}. Recent works have also shown that jointly optimizing both problems provides superior models that better capture the underlying structure of the target task~\cite{zelaautoml18,zimmertpami21aAPT}. 

Although the principle idea of applying AutoML to time series forecasting models is very natural, there are only few prior approaches addressing this~\cite{javeri2021improving,Li-kdd20AutoST,montero2020fforma,talagala2018meta}. In fact, combining state-of-the-art AutoML methods, such as Bayesian Optimization with multi-fidelity optimization~\cite{FalknerICML18BOHB,JamiesonAISTA16SH,klein-arxiv20a,LiJMLR17Hyperband}, with state-of-the-art time series forecasting models leads to several challenges we address in this paper. First, recent approaches for NAS mainly cover cell search spaces, allowing only for a very limited design space, that does not support different macro designs~\cite{dong-iclr20a,ying19NASBench101}. Our goal is to search over a large variety of different architectures covering state-of-the-art ideas. Second, evaluating DL models for time series forecasting is fairly expensive and a machine learning practicioner may not be able to afford many model evaluations. Multi-fidelity optimization, e.g.~\cite{LiJMLR17Hyperband}, was proposed to alleviate this problem by only allocating a fraction of the resources to evaluated configurations and promoting the most promising configurations to give them additional resources. Third, as a consequence of applying multi-fidelity optimization, we have to choose how different fidelities are defined, i.e. what kind of budget is used. Examples for such \emph{budget types} are number of epochs, dataset size or time series length. Depending on the correlation between lower and highest fidelity, multi-fidelity optimization can boost the efficiency of AutoML greatly or even slow it down in the worst case. Since we are the first to consider multi-fidelity optimization for AutoML on time series forecasting, we studied the efficiency of different budget types across many datasets. Fourth, all of these need to be put together; to that effect, we propose a new open-source package for Automated Deep Learning (AutoDL) for time series forecasting, dubbed \system.\footnote{The code is available under \url{https://github.com/automl/Auto-PyTorch}.}
Specifically, our contributions are as follows:
\begin{enumerate}
    \item We propose the AutoDL framework \system{} that is able to jointly optimize the architecture and the corresponding hyperparameters for a given dataset for time series forecasting.
    \item We present a unified architecture configuration space that contains several state-of-the-art forecasting architectures, allowing for a flexible and powerful macro-search.
    \item We provide insights into the configuration space of \system{} by studying the most important design decisions and show that different architectures are reasonable for different datasets.
    \item We show that \system{} is able to outperform a set of well-known traditional statistical models and modern deep learning models with an average relative error reduction of $19\%$ against the best baseline across many forecasting datasets.
\end{enumerate}

\section{Related Work}

We start by discussing the most closely related work in DL for time series forecasting, AutoDL, and AutoML for time series forecasting.

\subsection{Deep Learning based Forecasting}
Early work on forecasting focused on building a local model for each individual series to predict future values, ignoring the correlation between different series. 
In contrast, global forecasting models are able to capture information of multiple time series in a dataset and use this at prediction time~\cite{januschowski2020criteria}.
With growing dataset size and availability of multiple time series from similar sources, this becomes increasingly appealing over local models. 
We will in the following briefly introduce some popular forecasting DL models.

Simple feed-forward MLPs have been used for time series forecasting and extended to more complex models. For example, the N-BEATS framework~\cite{OreshkinICLR2020NBEATS} is composed of multiple stacks, each consisting of several blocks.
This architectural choice aligns with the main principle of modern architecture design: Networks should be designed in a block-wise manner instead of layer-wise~\cite{Zoph18NAS}.

Additionally, RNNs~\cite{ChoEMNLP14Seq2Seq,Schmidhuber1997LSTM} were proposed to process sequential data and thus they are directly applicable to time series forecasting~\cite{HEWAMALAGE2021RNN,wen2017multi}. A typical RNN-based model is the Seq2Seq network~\cite{ChoEMNLP14Seq2Seq} that contains an RNN encoder and decoder. Wen et al.~\cite{wen2017multi} further replaced the Seq2Seq's RNN decoder with a multi-head MLP. Flunkert et al.~\cite{Flunkert17DeepAR} proposed DeepAR that wraps an RNN encoder as an auto-regressive model and uses it to iteratively generate new sample points based on sampled trajectories from the last time step.

In contrast, CNNs can extract local, spatially-invariant relationships. Similarly, time series data may have time-invariant relationships, which makes CNN-based models suitable for time series tasks, e.g. WaveNet~\cite{borovykh2017WaveNet,OordISCA16Wavenet} and Temporal Convolution Networks (TCN)~\cite{Bai18TCN}. Similar to RNNs, CNNs could also be wrapped by an auto-regressive model to recursively forecast future targets~\cite{borovykh2017WaveNet,OordISCA16Wavenet}.
 
Last but not least, attention mechanisms and transformers have shown superior performance over RNNs on natural language processing tasks~\cite{VaswaniNIPS17Transformer} and over CNNs on computer vision tasks~\cite{dosovitskiy2021VIT}.
Transformers and RNNs can also be combined; e.g. Lim et al.~\cite{lim2019tft} proposed temporal fusion transformers (TFT) that stack a transformer layer on top of an RNN to combine the best of two worlds.

\subsection{Automated Deep Learning (AutoDL)}
State-of-the-art AutoML approaches include Bayesian Optimization (BO)~\cite{FeurerNIPS15ASKL}, Evolutionary Algorithms (EA)~\cite{Olson16TOPT}, reinforcement learning~\cite{Zoph18NAS} or ensembles~\cite{Erickson20AutoGluon}. Most of them consider AutoML system as a black-box optimization problem that aims at finding the most promising machine learning models and their optimal corresponding hyperparameters. Neural Architecture Search (NAS), on the other hand, only contains one search space: its architecture. NAS aims at finding the optimal architecture for the given task with a fixed set of hyperparameters. Similar to the traditional approach, the architecture could be optimized with BO~\cite{jin2019autokeras,zimmertpami21aAPT}, EA~\cite{Real19RENAS} or Reinforcement Learning~\cite{Zoph18NAS} among others, but there also exist many NAS-specific speedup techniques, such as one-shot models~\cite{Xiao_2020} and zero-cost proxies~\cite{abdelfattah2021zerocost}. In this work we follow the state-of-the-art approach from Auto-PyTorch~\cite{zimmertpami21aAPT} and search for both the optimal architecture and its hyperparameters with BO.

Training a deep neural network requires lots of computational resources. Multi-fidelity optimization~\cite{FalknerICML18BOHB,JamiesonAISTA16SH,LiJMLR17Hyperband} is a common approach to accelerate AutoML and AutoDL. 
It prevents the optimizer from investing too many resources on the poorly performing configurations and allows for spending more on the most promising ones. 
However, the correlation between different fidelities might be weak~\cite{ying19NASBench101} for DL models, in which case
the result on a lower fidelity will provide little information for those on higher fidelities. Thus, it is an open question how to properly select the budget type for a given target task, and researchers often revert to application-specific decisions. 

\subsection{AutoML for Time Series Forecasting}
While automatic forecasting has been of interest in the research community in the past~\cite{hyndman2008automatic}, dedicated AutoML approaches for time series forecasting problems have only been explored recently \cite{halvari2021robustness,javeri2021improving,kurian2021boat,meisenbacher2022review,shah2021autoai-ts}. 
Optimization methods such as random search~\cite{vanKuppeveltSoftwareX2020}, genetic algorithms \cite{dahl2020tspo}, monte carlo tree search and algorithms akin to multi-fidelity optimization \cite{shah2021autoai-ts} have been used among others.
Paldino et al.~\cite{paldino2021does} showed that AutoML frameworks not intended for time series forecasting originally - in combination with feature engineering - were not able to significantly outperform simple forecasting strategies; a similar approach is presented in \cite{dahl2020tspo}.
As part of a review of AutoML for forecasting pipelines, Meisenbacher et al.~\cite{meisenbacher2022review} concluded that there is a need for optimizing the entire pipeline as existing works tend to only focus on certain parts.
We took all of these into account by proposing \system{} as a framework that is specifically designed to optimize over a flexible and powerful configuration space of forecasting pipelines.

\section{AutoPyTorch Forecasting}
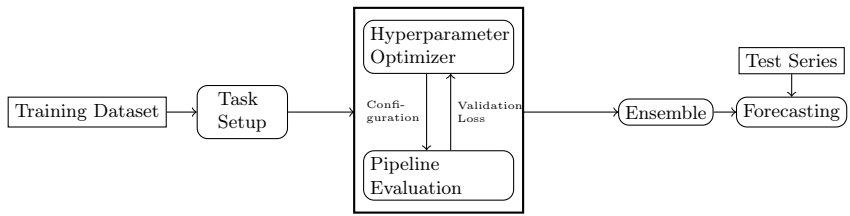
\begin{figure}[t]
    \centering
    \begin{tikzpicture}
    [scale=0.8,
    every node/.style={scale=0.8},block/.style={rectangle, minimum width=15mm, minimum height=4mm, draw=black, rounded corners, align=left},
    module/.style={rectangle, minimum width=25mm, minimum height=4mm, draw=black, align=left},
    data/.style={rectangle, minimum width =15mm, minimum height=4mm, draw=black}]
    
    \node (dataset) [data] {Training Dataset};

    \node (tasksetup) [block,  right = 4mm of dataset] {Task \\Setup};

    \node (hpoptimizer) [block, above right = 1.5mm and 10mm of tasksetup, text width=7em] {Hyperparameter\\Optimizer};
    \node (pipeline) [block, below right =1.5mm and 10mm of tasksetup, text width=7em] {Pipeline\\ Evaluation};
    
    \begin{scope}[local bounding box=hpo]
    \draw[thick] ($(pipeline.south west)+(-0.15, -0.2)$) rectangle ($(hpoptimizer.north east)+(0.15,0.2)$);
  \end{scope}

    \node (ensemble) [block, right = 60mm of dataset] {Ensemble};
    \node (forecasting) [block, right = 3mm of ensemble] {Forecasting};
    \node (test) [data, above = 3mm of forecasting] {Test Series};
    
    \draw[->] (dataset.east) -- (tasksetup.west);
    \draw[->] (test.south) -- (forecasting.north);
    
     \draw[->] ([xshift=-2mm]hpoptimizer.south) -- ([xshift=-2mm]pipeline.north) node[font =\tiny, midway, left, align=left] {Confi-\\guration};
          \draw[->] ([xshift=2mm]pipeline.north) -- ([xshift=2mm]hpoptimizer.south) node[font =\tiny, midway, right, align=left] {Validation\\Loss};
    
    \draw[->] (tasksetup.east) -- (tasksetup.east -| hpo.west);
    
    \draw[->] (hpo.east |- ensemble.west) -- (ensemble.west);
    
    \draw[->] (ensemble.east) -- (forecasting.west);

    \end{tikzpicture}
    \caption{An overview of \system. Given a dataset, \system{} automatically prepares the data to fit the requirement of a forecasting pipeline. The AutoML optimizer will then use the selected budget type to search for desirable neural architectures and hyperparameters from the pipeline configuration space. Finally, we create an ensemble out of the most promising pipelines to do the final forecasting on the test sets.}
    \label{fig:aptsystem}
\end{figure}

For designing an AutoML system, we need to consider the following components: optimization targets, configuration space and optimization algorithm. The high-level workflow of our \system{} framework is shown in Figure~\ref{fig:aptsystem}; in many ways it functions similar to existing state-of-the-art AutoML frameworks~\cite{feurer2019Hyperparameter,zimmertpami21aAPT}. To better be able to explain unique design choice for time series forecasting, we first present a formal statement of the forecasting problem and discuss challenges in evaluating forecasting pipelines before describing the components in detail.

\subsection{Problem Definition}
A multi-series forecasting task is defined as follows: given a dataset that contains $\numseq$ series: $\dset = \{\dsetseq\}_{\seqidx=1}^\numseq$ and $\dsetseq$ represents one series in the dataset: $\dsetseq = \{\targetseq{\observedintervalseq}, \featurespastseq{\observedintervalseq}, \featuresfutureseq{\forecastintervalseq}\}$\footnote{For the sake of brevity, we omit the sequence index $i$ in the following part of this paper unless stated otherwise.}, 
where $\seqlength$ is the number of time steps until forecasting starts; $\horizon$ is the forecasting horizon that the model is required to predict; $\target_{\observedinterval}$,  $\featurespast_{\observedinterval}$ and $\featuresfuture_{\forecastintervalseq}$ are the sets of observed past targets, past features and known future features values, respectively. The task of time series forecasting is to predict the possible future values with a model trained on $\dset$:
\begin{equation}
    \prediction_{\forecastinterval} = f(\target_{\observedinterval}, \feature_{\allinterval}; \weights)
\end{equation}
where $\feature_{\allinterval} := [\featurespast_{\observedinterval}, \featuresfuture_{\forecastinterval}]$, $\weights$ are the model parameters that are optimized with training losses $\loss_{train}$, and $\prediction_{\forecastinterval}$ are the predicted future target values. Depending on the model type,  $\prediction_{\forecastinterval}$ can be distributions~\cite{Flunkert17DeepAR} or scalar values~\cite{OreshkinICLR2020NBEATS}. Finally, the forecasting quality is measured by the discrepancy between the predicted targets $\prediction_{\forecastinterval}$ and the ground truth future targets $\target_{\forecastinterval}$ according to a defined loss function $\loss$. The most commonly applied metrics include mean absolute scaled error (MASE), mean absolute percentage error (MAPE), symmetric mean absolute percentage error (sMAPE) and mean absolute error (MAE)~\cite{FLORES198693SMAPE,HYNDMAN2006679MASE,OreshkinICLR2020NBEATS}.

\subsection{Evaluating Forecasting Pipelines\label{sec:tasksetting}}

We split each sequence into three parts to obtain: a training set $\dsettrain=\{\target_{\traininterval}, \feature_{\allinterval}\}$,
a validation set $\dsetval=\{\target_{\valinterval}, \feature_{\valinterval}\}$ and a test set $\dsettest=\{\target_{\testinterval}, \feature_{\testinterval}\}$, i.e., the tails of each sequences are reserved as $\dsetval$.
At each iteration, our AutoML optimizer suggests a new hyperparameter and architecture configuration $\conf$, trains it on $\dsettrain$ and evaluates it on $\dsetval$. 

Both in AutoML frameworks~\cite{FeurerNIPS15ASKL,zimmertpami21aAPT} and in forecasting frameworks~\cite{OreshkinICLR2020NBEATS}, ensembling of  models is a common approach. We combine these two worlds in \system{} by using ensemble selection~\cite{caruana-icml04a} to construct a weighted ensemble that is composed of the best $k$ forecasting models from the previously evaluated configurations $\hist$. Finally, we retrain all ensemble members on $\dsetval \cup \dsettrain$ before evaluating on $\dsettest$. 

\subsection{Forecasting Pipeline Configuration Space \label{pipeline}}

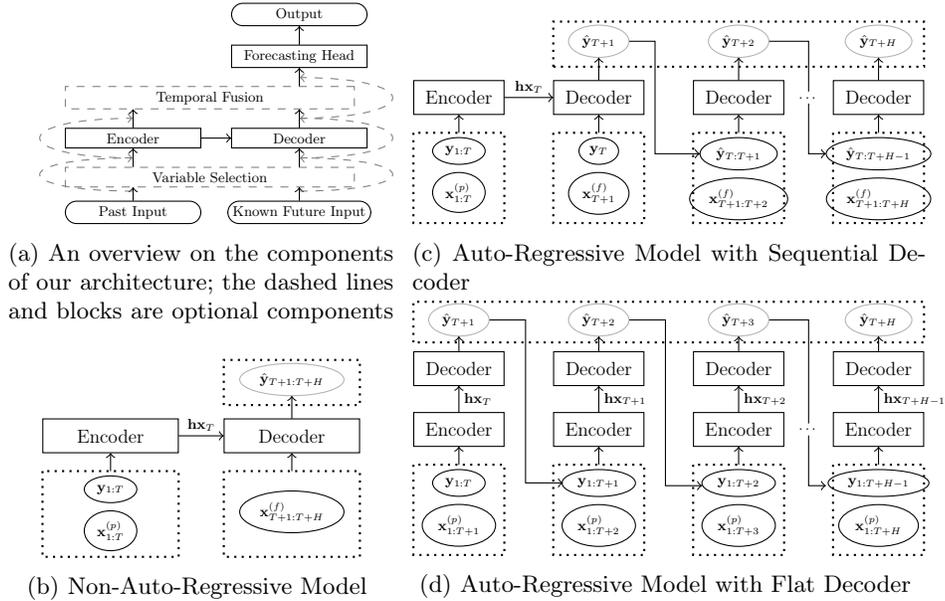
\begin{figure}[t]
\centering
\begin{subfigure}[t]{0.42\textwidth}
\centering
    \begin{tikzpicture}[
    scale=0.6,
every node/.style={scale=0.6},
    block/.style={rectangle, minimum width=30mm, minimum height=4mm, draw=black},
    globalblock/.style={rectangle, minimum width=64mm, minimum height=4mm, draw=gray, dashed},
    data/.style={rectangle, minimum width=30mm, minimum height=4.5mm, draw=black, rounded corners=.15cm},
    dot/.style = {circle, fill, minimum size=3pt,inner sep = 0, outer sep=0pt,node contents={}},
    ]
    
    \node[block] (encoder) {Encoder};
    \node[block, right=4mm of encoder] (decoder) {Decoder};
    
    \node[globalblock, below =2.5mm  of encoder.south west, anchor=north west] (variableselection) {Variable Selection};

    \node[data, below=7mm of encoder] (inputpast) {Past Input};
    
    \node[data, below=7mm of decoder] (inputfuture) {Known Future Input};
    
    \node[globalblock, above = 2.5mm of encoder.north west, anchor=south west]
    (temporalfusion) {Temporal Fusion};
    
    \node[block, above=8mm of decoder.north]
    (head) {Forecasting Head};
    
    \node[data, above=2.5mm of head]
    (output) {Output};
    
     \node[inner sep=0,minimum size=0,above =1.25mm of inputpast] (skipencoder1) {};
     
     \node[inner sep=0,minimum size=0,below =1.25mm of encoder] (skipencoder2) {}; 
     \node[inner sep=0,minimum size=0,above =1.25mm of encoder] (skipencoder3) {}; 
     
    \node[inner sep=0,minimum size=0,above =1.25mm of inputfuture] (skipdecoder1) {};
     
     \node[inner sep=0,minimum size=0,below =1.25mm of decoder] (skipdecoder2) {}; 
     \node[inner sep=0,minimum size=0,above =1.25mm of decoder] (skipdecoder3) {}; 
     
    \node[inner sep=0,minimum size=0,below =1.25mm of head] (skipdecoder4) {}; 
    
    \draw[->] (inputpast.north) -- (inputpast.north |- variableselection.south);
    
    \draw[->] (variableselection.north -| encoder.south) -- (encoder.south);
    
    \draw[->] (encoder.north) -- (encoder.north |- temporalfusion.south);
    
    \draw[->] (inputfuture.north) -- (inputfuture.north |- variableselection.south);
    
    \draw[->] (variableselection.north -| decoder.south) -- (decoder.south);
    
    \draw[->] (decoder.north) -- (decoder.north |- temporalfusion.south);
    
    \draw[->] (encoder.east) -- (decoder.west);
    
    \draw[->] (temporalfusion.north -| head.south) -- (head.south);
    
    \draw[->] (head.north) -- (output.south);
    
    \draw[->, gray, dashed] (skipencoder1) .. controls ++(-27.5mm, 0.1mm) and ++(-27.5mm, -0.1mm) .. (skipencoder2); 
    
    \draw[->, gray, dashed] (skipencoder2) .. controls ++(-27.5mm, 0.1mm) and ++(-27.5mm, -0.1mm) .. (skipencoder3);
    
    \draw[->, gray, dashed] (skipdecoder1) .. controls ++(27.5mm, 0.1mm) and ++(27.5mm, -0.1mm) .. (skipdecoder2); 
    
    \draw[->, gray, dashed] (skipdecoder2) .. controls ++(27.5mm, 0.1mm) and ++(27.5mm, -0.1mm) .. (skipdecoder3);

    \draw[->, gray, dashed] (skipdecoder3) .. controls ++(27.5mm, 0.1mm) and ++(27.5mm, -0.1mm) .. (skipdecoder4);

    \end{tikzpicture}
    
    \caption{An overview on the components of our architecture; the dashed lines and blocks are optional components \label{fig:architectureoverview}}
\begin{tikzpicture}[
scale=0.6,
every node/.style={scale=0.6},
features/.style={shape=ellipse, draw=black,minimum size=5mm},
encoder/.style={rectangle,   minimum width=30mm, minimum height=7.5mm, draw=black, font=\large},
decoder/.style={rectangle,   minimum width=30mm, minimum height=7.5mm, draw=black, font=\large},
targets/.style={shape=ellipse,  draw=black, minimum size=5mm},
netoutput/.style={shape=ellipse, draw=black!30, minimum size=5mm},
]

\node[encoder] (encoder) {Encoder};

\node[targets, below = 3mm of encoder] (targetpast) {$\target_{\observedinterval}$};
\node[features, below = 1mm of targetpast] (featurespast) {$\featurespast_{\observedinterval}$};
\draw[thick,dotted]     ($(encoder.south west)+(-0.0, -0.4)$) rectangle ($(encoder.south east)+(0.0,-2.3)$);
\draw[->] ($(encoder.south)+(0.0, -0.4)$) -- (encoder.south);

\node[decoder, right = 6mm of encoder] (decoder) {Decoder};
\draw[->] (encoder.east) -- (decoder.west) node[midway, above] {$\hiddenstate_\seqlength$};

\node[features, below = 5mm of decoder] (featurespast) {$\featuresfuture_{\forecastinterval}$};
\draw[thick,dotted]     ($(decoder.south west)+(-0.0, -0.4)$) rectangle ($(decoder.south east)+(0.0,-2.3)$);
\draw[->] ($(decoder.south)+(0.0, -0.4)$) -- (decoder.south);

\node[netoutput, above = 3mm of decoder.north] (outputs) {$\prediction_{\forecastinterval}$};
\draw[->] (decoder.north) -- (outputs.south);
\draw[thick, dotted] ($(decoder.north west)+(-0.0, 0.3)$) rectangle ($(decoder.north east)+(0.0,1.3)$);
\end{tikzpicture}

\caption{Non-Auto-Regressive Model \label{fig:forwardnetwork}}
\end{subfigure}
\hfill
\begin{subfigure}[t]{0.56\textwidth}
\centering
\begin{tikzpicture}[
scale=0.6,
every node/.style={scale=0.6},
features/.style={shape=ellipse, draw=black,minimum size=5mm},
encoder/.style={rectangle,   minimum width=20mm, minimum height=7.5mm, draw=black, font=\large},
decoder/.style={rectangle,   minimum width=20mm, minimum height=7.5mm, draw=black, font=\large},
targets/.style={shape=ellipse,  draw=black, minimum size=5mm},
netoutput/.style={shape=ellipse, draw=black!30, minimum size=5mm},
]
\node[encoder] (encoder) {Encoder};

\node[targets, below = 3mm of encoder] (targetpast) {$\target_{\observedinterval}$};
\node[features, below = 1mm of targetpast] (featurespast) {$\featurespast_{\observedinterval}$};
\draw[thick,dotted]     ($(encoder.south west)+(-0.0, -0.4)$) rectangle ($(encoder.south east)+(0.0,-2.4)$);
\draw[->] ($(encoder.south)+(0.0, -0.4)$) -- (encoder.south);

\node[decoder, right = 6.5mm of encoder] (decoder1) {Decoder};
\draw[->] (encoder.east) -- (decoder1.west) node[midway, above] {$\hiddenstate_\seqlength$};

\node[targets, below = 3mm of decoder1] (target1) {$\target_{\seqlength}$};
\node[features, below = 1mm of target1] (feature1) {$\featuresfuture_{\seqlength+1}$};
\draw[thick,dotted]     ($(decoder1.south west)+(-0.0, -0.4)$) rectangle ($(decoder1.south east)+(0.0,-2.4)$);
\draw[->] ($(decoder1.south)+(0.0, -0.4)$) -- (decoder1.south);

\node[netoutput, above = 3mm of decoder1.north] (outputs1) {$\prediction_{\seqlength + 1}$};
\draw[->] (decoder1.north) -- (outputs1.south);

\node[decoder, right = 6.5mm of decoder1] (decoder2) {Decoder};

\node[targets, below = 3mm of decoder2] (target2) {$\prediction_{\seqlength: \seqlength + 1}$};
\node[features, below = 1mm of target2] (feature2) {$\featuresfuture_{\seqlength + 1: \seqlength+2}$};
\draw[thick,dotted] ($(decoder2.south west)+(-0.0, -0.4)$) rectangle ($(decoder2.south east)+(0.0,-2.4)$);
\draw[->] ($(decoder2.south)+(0.0, -0.4)$) -- (decoder2.south);
\node[netoutput, above = 3mm of decoder2.north] (outputs2) {$\prediction_{\seqlength + 2}$};
\draw[->] (decoder2.north) -- (outputs2.south);

\draw[->] (outputs1.east) -- ++(0.57, 0) |- (target2.west);

\node[decoder, right = 6.5mm of decoder2] (decoderH) {Decoder};
\node at ($(decoder2)!.5!(decoderH)$) (dots_node) {\ldots};

\draw (outputs2.east) -| (dots_node.north);

\node[targets, below = 3mm of decoderH] (targetH) {$\prediction_{\seqlength: \seqlength + H - 1}$};

\draw[->] (dots_node.south) |- (targetH.west);

\node[features, below = 1mm of targetH] (featureH) {$\featuresfuture_{\seqlength + 1:\seqlength+H}$};
\draw[thick,dotted] ($(decoderH.south west)+(-0.0, -0.4)$) rectangle ($(decoderH.south east)+(0.0,-2.4)$);
\draw[->] ($(decoderH.south)+(0.0, -0.4)$) -- (decoderH.south);
\node[netoutput, above = 3mm of decoderH.north] (outputsH) {$\prediction_{\seqlength + H}$};
\draw[->] (decoderH.north) -- (outputsH.south);

\draw[thick, dotted] ($(decoder1.north west)+(-0.0, 0.3)$) rectangle ($(decoderH.north east)+(0.0,1.3)$);

\end{tikzpicture}

\caption{Auto-Regressive Model with Sequential Decoder \label{fig:seq2seq}}

\begin{tikzpicture}[
scale=0.6,
every node/.style={scale=0.6},
features/.style={shape=ellipse, draw=black,minimum size=5mm},
encoder/.style={rectangle,   minimum width=20mm, minimum height=7.5mm, draw=black, font=\large},
decoder/.style={rectangle,   minimum width=20mm, minimum height=7.5mm, draw=black, font=\large},
targets/.style={shape=ellipse,  draw=black, minimum size=5mm},
netoutput/.style={shape=ellipse, draw=black!30, minimum size=5mm},
]

\node[encoder,] (encoder0) {Encoder};
\node[encoder, right = 6.5mm of encoder0] (encoder1) {Encoder};
\node[encoder, right = 6.5mm of encoder1] (encoder2) {Encoder};
\node[encoder, right = 6.5mm of encoder2] (encoder3) {Encoder};

\foreach \t [count=\ti from 1]in {0, 1, 2, 3}
{

\node[targets, below = 3mm of encoder\t] (targetpast\t) {$\target_{\observedinterval{\ifthenelse{\t=0 }{}{\ifthenelse{\t=3}{+H-1}{+\t}}}}$};
\node[features, below = 1mm of targetpast\t] (featurespast\t) {$\featurespast_{\observedinterval+\ifthenelse{\t=3}{H}{\ti}}$};
\draw[thick,dotted]     ($(encoder\t.south west)+(-0.0, -0.4)$) rectangle ($(encoder\t.south east)+(0.0,-2.4)$);
\draw[->] ($(encoder\t.south)+(0.0, -0.4)$) -- (encoder\t.south);

\node[decoder, above = 3.5mm of encoder\t.north] (decoder\t) {Decoder};
\draw[->] (encoder\t.north) -- (decoder\t.south) node[midway, right] {$\hiddenstate_{\seqlength\ifthenelse{\t=0}{}{+\ifthenelse{\t=3}{H-1}{\t}}}$};

\node[netoutput, above = 2mm of decoder\t.north] (outputs\t) {$\prediction_{\seqlength + \ifthenelse{\t=3}{H}{\ti}}$};
\draw[->] (decoder\t.north) -- (outputs\t.south);
}

\draw[->](outputs0.east)  -- ++(0.8, 0) |- ($(target1.west) + (-0.35, 0)$);
\draw[->](outputs1.east)  -- ++(0.8, 0) |- ($(target2.west) + (0.05, 0)$);

\node at ($(encoder2)!.5!(encoder3)$) (dots_node) {\ldots};
\draw (outputs2.east) -| (dots_node.north);
\draw[->] (dots_node.south) |- (targetH.west);

\draw[thick, dotted] ($(decoder0.north west)+(-0.0, 0.2)$) rectangle ($(decoder3.north east)+(0.0,1.1)$);

\end{tikzpicture}

\caption{Auto-Regressive Model with Flat Decoder  \label{fig:deepAR}}

\end{subfigure}

\caption{Overview of the architectures that can be built by our framework. (a) shows the main components of our architecture space. (b)-(d) are specific instances of (a)  and its data flow given different architecture properties. \label{fig:architectures}}
\end{figure}
Existing DL packages for time series forecasting~\cite{gluonts_jmlr,beitner2020pytorchforecasting} follow the typical structure of traditional machine learning libraries: models are built individually with their own hyperparameters. Similar to other established AutoML tools~\cite{Erickson20AutoGluon,FeurerNIPS15ASKL,Olson16TOPT}, we designed the configuration space of \system{} as a combined algorithm selection and hyperparameter (CASH) problem~\cite{ThorntonKDD13AutoWEKA}, i.e.,  the optimizer first selects the most promising algorithms and then optimizes for their optimal hyperparameter configurations, with a hierarchy of design decisions. Deep neural networks, however, are built with stacked blocks~\cite{Zoph18NAS} that can be disentangled to fit different requirements~\cite{Wu21FBNetV5}. For instance, Seq2Seq~\cite{ChoEMNLP14Seq2Seq},  MQ-RNN~\cite{wen2017multi} and DeepAR~\cite{Flunkert17DeepAR} all contain an RNN as their encoders. These models naturally share common aspects and cannot be simply treated as completely different models. To fully utilize the relationships of different models, we propose a configuration space that includes all the possible components in a forecasting network.

As shown in Figure~\ref{fig:architectureoverview}, most existing forecasting architectures can be decomposed into 3 parts: \textit{encoder}, \textit{decoder} and \textit{forecasting heads}: the encoder receives the past target values and embeds them into the latent space. The latent embedding, together with the known future features (if applicable), are fed to the decoder network; the output of the decoder network is finally passed to the forecasting head to generate a sequence of scalar values or distributions, depending on the type of forecasting head. Additionally, the \textit{variable selection}, \textit{temporal fusion} and \textit{skip connection layers} introduced by TFT~\cite{lim2019tft} can be seamlessly integrated into our networks and are treated as optional components. 
\begin{table}[t]
\resizebox{\textwidth}{!}{
\begin{tabular}{c|c|c|c|c  }
 \toprule
  \multicolumn{2}{c|}{Encoder} & Decoder& auto-regressive  & Architecture Class \\
  
   \midrule

 \multirow{2}{*}{Flat Encoder} & MLP & MLP & No & Feed Forward Network \\

   \cline{2-5} & N-BEATS & N-BEATS & No  & N-BEATS \cite{OreshkinICLR2020NBEATS}\\
 \hline
 \multirow{6}{*}{Seq. Encoder} & \multirow{4}{*}{RNN/Transformer} & \multirow{2}{*}{RNN/Transformer} & Yes & Seq2Seq~\cite{ChoEMNLP14Seq2Seq} \\
  \cline{4-5}&  & & No & TFT~\cite{lim2019tft} \\
  
  \cline{3-5} & & \multirow{2}{*}{MLP} & Yes  & DeepAR \cite{Flunkert17DeepAR} \\
  
  \cline{4-5} & & & No & MQ-RNN \cite{wen2017multi} \\
    
  \cline{2-5} & \multirow{2}{*}{TCN}& \multirow{2}{*}{MLP} & Yes &  DeepAR~\cite{Flunkert17DeepAR}/WaveNet~\cite{OordISCA16Wavenet} \\
    
    \cline{4-5} & &  & No  & MQ-CNN ~\cite{wen2017multi}  \\
    \bottomrule
\end{tabular}
}
\caption{An overview of the possible combinations and design decisions of the models that exists in our configuration space. Only the TFT Network contains the optional components presented in Figure~\ref{fig:architectureoverview}. \label{table:archtecture}}
\end{table}

Table \ref{table:archtecture} lists all possible choices of encoders, decoders, and their corresponding architectures in our configuration space. Specifically, we define two types of network components: sequential encoder (Seq. Encoder) and flat encoder (Flat Encoder). The former (e.g., RNN, Transformer and TCN) directly processes sequential data and output a new sequence; the latter (e.g., MLP and N-BEATS) needs to flatten the sequential data into a 2D matrix to fuse the information from different time steps.
Through this configuration space, \system{} is able to encompass the ``convex hull'' of several state-of-the-art global forecasting models and tune them.

As shown in Figure \ref{fig:architectures}, given the properties of encoders, decoders, and models themselves, we construct three types of architectures that forecast the future targets in different ways. Non-Auto-Regressive models (Figure \ref{fig:forwardnetwork}), including MLP, MQ-RNN, MQ-CNN, N-BEATS and TFT, forecast the multi-horizontal predictions within one single step. In contrast, Auto-Regressive models do only one-step forecasting within each forward pass. The generated forecasting values are then iteratively fed to the network to forecast the value at the next time step. All the auto-regressive models are trained with teacher forcing~\cite{HEWAMALAGE2021RNN}. Only sequential networks could serve as an encoder in auto-regressive models, however, we could select both sequential and flat decoders for auto-regressive models. Sequential decoders are capable of independently receiving the newly generated predictions. We consider this class of architectures as a Seq2Seq~\cite{ChoEMNLP14Seq2Seq} model: we first feed the past input values to the encoder to generate its output $\hiddenstate$ and then pass $\hiddenstate$ to the decoder, as shown in Figure~\ref{fig:seq2seq}. Having acquired $\hiddenstate$, the decoder then generates a sequence of predictions with the generated predictions and known future values by itself. Finally, Auto-Regressive Models with flat decoders are classified as the family of DeepAR models~\cite{Flunkert17DeepAR}. As the decoder could not collect more information as the number of generated samples increases, we need to feed the generated samples back to the encoder, as shown in Figure~\ref{fig:deepAR}. 

Besides its architectures, hyperparemters also play an important role on the performance of a deep neural network~\cite{zelaautoml18}, for the details of other hyperparameters in our configuration space, we refer to the Appendix.
\subsection{Hyperparameter Optimization \label{sec:optimization}}
We optimize the loss on the validation set $\loss_{\dsetval}$ with BO~\cite{feurer2019Hyperparameter}. It is known for its sample efficiency, making it a good approach for expensive black-box  optimization tasks, such as AutoDL for expensive global forecasting DL models. Specifically, we optimize the hyperparameters with SMAC~\cite{Hutter2011SMAC}\footnote{We used SMAC3~\cite{lindauer2021smac3} from \url{https://github.com/automl/SMAC3}} that constructs a random forest to model the loss distribution over the configuration space. 

Similar to other AutoML tools~\cite{FeurerNIPS15ASKL,zimmertpami21aAPT} for supervised classification, we utilize multi-fidelity optimization to achieve better any-time performance. Multi-fidelity optimizers start with the lowest budget and gradually assign higher budgets to well-performing configurations. Thereby, the choice of what budget type to use is essential for the efficiency of a multi-fidelity optimizer. 
The most popular choices of budget type in DL tasks are the number of epochs and dataset size. For time series forecasting, we propose the following four different types of budget:
\begin{itemize}
    \item Number of Epochs (\textit{$\#$Epochs})
    \item Series Resolution (\textit{Resolution})
    \item Number of Series (\textit{$\#$Series})
    \item Number of Samples in each Series (\textit{$\#$SMPs per Ser.})
\end{itemize}

A higher \textit{Resolution} indicates an extended sample interval. The sample interval is computed by the inverse of the fidelity value, e.g., a resolution fidelity of $0.1$ indicates for each series we take every tenth point: we shrink the size of the sliding window accordingly to ensure that the lower fidelity optimizer does not receive more information than the higher fidelity optimizer. \textit{$\#$Series} means that we only sample a fraction of sequences to train our model. Finally, \textit{$\#$SMPs per Ser.} indicates that we decrease the expected value of the number of samples within each sequence; see Section \ref{sec:tasksetting} for sample-generation method. Next to these multi-fidelity variants, we also consider vanilla Bayesian optimization (\textit{Vanilla BO}) using the maximum of all these fidelities.

\subsection{Proxy-Evaluation on Many Time Series\label{sec:prox-eval}}

All trained models must query every series to evaluate $\loss_{val}$. However, the number of series could be quite large. Additionally, many forecasting models (e.g., DeepAR) are cheap to be trained but expensive during inference time. As a result, rather than training time, inference time is more likely to become a bottleneck to optimize the hyperparameters on a large dataset (for instance, with $10$k series or more), where configuration with lower fidelities would no longer provide the desirable speed-up when using the full validation set. Thereby, we consider a different evaluation strategy on large datasets (with more than $1$k series) and lower budgets: we ask the model to only evaluate a fraction of the validation set (we call this fraction ``proxy validation set'') while the other series are predicted by a dummy forecaster (which simply repeats the last target value in the training series, i.e., $\target_{\seqlength}$, $\horizon$ times). The size of the proxy validation set is proportional to the budget allocated to the configuration: maximal budget indicates that the model needs to evaluate the entire validation set. We set the minimal number of series in the proxy set to be $1$k to ensure that it contains enough information from the validation set. The proxy validation set is generated with a grid to ensure that all the configurations under the same fidelity are evaluated on the same proxy set.  

\section{Experiments}
We evaluate \system{} on the established benchmarks of the Monash Time Series Forecasting Repository~\cite{godahewa2021monash}\footnote{\url{https://forecastingdata.org/}}. This repository contains various datasets that come from different domains, which allows us to assess the robustness of our framework against different data distributions. Additionally, it records the performance of several models, including local models~\cite{assimakopoulos2000theta,box2015timeARIMA,delivera2011TBATS,hyndman2008forecastingETS,hyndman2021forecasting}, global traditional machine learning models ~\cite{ProkhorenkovaNeurIPS18CatBoost,Trapero2015PR}, and global DL models~\cite{gluonts_jmlr,borovykh2017WaveNet,OreshkinICLR2020NBEATS,Flunkert17DeepAR,VaswaniNIPS17Transformer} on $\dsettest$,  see \cite{godahewa2021monash} for details. For evaluating \system{}, we will follow the exact same protocol and dataset splits. We focus our comparison of \system{} against two types of baselines: (i) the overall single best baseline from \cite{godahewa2021monash}, assuming a user would have the required expert knowledge and (ii) the best dataset-specific baseline. We note that the latter is a very strong baseline and a priori it is not known which baseline would be best for a given dataset; thus we call it the \textit{theoretical oracle baseline}.
Since the Monash Time Series Forecasting Repository does not record the standard deviation of each method, we reran those baselines on our cluster for 5 times. Compared to the repository, our configuration space includes one more strong class of algorithms, TFT~\cite{lim2019tft}, which we added to our set of baselines to ensure a fair and even harder comparison.

We set up our task following the method described in Section~\ref{sec:tasksetting}: HPO is only executed on $\dset_{train/val}$  while $\horizon$ is given by the original repository. As described in Section~\ref{sec:tasksetting}, we create an ensemble with size $20$ that collects multiple models during the course of optimization. When the search finishes, we refit the ensemble to the union of $\dset_{train/val}$ and evaluate the refitted model on $\dsettest$. Both $\loss_{val}$ and $\loss_{test}$ are measured with the mean value of MASE~\cite{HYNDMAN2006679MASE} across all the series in the dataset. To leverage available expert knowledge, \system{} runs an initial design with the default configurations of each model in Table~\ref{table:archtecture}. Please note that this initial design will be evaluated on the smallest available fidelity. 
All multi-fidelity variants of \system{} start with the cheapest fidelity of $1/9$, use then $1/3$ and end with the highest fidelity ($1.0$).
The runs of \system{} are repeated $5$ times with different random seeds.

We ran all the datasets on a cluster node equipped with 8 Intel Xeon Gold 6254@ 3.10GHz CPU cores and one NVIDIA GTX 2080TI GPU equipped with PyTorch 1.10 and Cuda 11.6. The hyperparameters were optimized with SMAC3 v1.0.1 for $10$ hours, and then we refit the ensemble on $\dset_{train/val}$ and evaluate it on the test set. All the jobs were finished within $12$ hours.

\subsection{Time Series Forecasting \label{sec:forecast_eval}}

\begin{table}[tbph]
\resizebox{\textwidth}{!}{

\begin{tabular}{l| rrrrr|rr}
\toprule
            & \multicolumn{5}{c|}{\system} & & \\ 
 &             $\#$~Epochs &    Resolution &             $\#$~Series &      $\#$~SMPs per Ser. &                     Vanilla BO &               \thead{Best dataset- \\ specific baseline} & \thead{Overall single \\ best baseline} \\
\midrule

M3 Yearly                  &           2.73(0.10) &  2.66(0.05) &           2.76(0.09) &  \textbf{2.64}(0.09) &           2.68(0.08) &           2.77(0.00) &  3.13(0.00) \\
M3 Quarterly               &  \textbf{1.08}(0.01) &  1.10(0.01) &           1.10(0.01) &           1.09(0.02) &           1.12(0.03) &           1.12(0.00) &  1.26(0.00) \\
M3 Monthly                 &  \textbf{0.85}(0.01) &  0.89(0.02) &           0.86(0.01) &           0.87(0.04) &           0.86(0.02) &           0.86(0.00) &  0.86(0.00) \\
M3 Other                   &           1.90(0.07) &  1.82(0.03) &           1.98(0.13) &           1.92(0.05) &           1.95(0.15) &  \textbf{1.81}(0.00) &  1.85(0.00) \\
M4 Quarterly               &           1.15(0.01) &  1.13(0.01) &  \textbf{1.13}(0.01) &           1.15(0.01) &           1.15(0.02) &           1.16(0.00) &  1.19(0.00) \\
M4 Monthly                 &           0.93(0.02) &  0.93(0.02) &           0.93(0.02) &  \textbf{0.93}(0.02) &           0.96(0.02) &           0.95(0.00) &  1.05(0.00) \\
M4 Weekly                  &           0.44(0.01) &  0.45(0.02) &  \textbf{0.43}(0.02) &           0.44(0.02) &           0.45(0.01) &           0.48(0.00) &  0.50(0.00) \\
M4 Daily                   &           1.14(0.01) &  1.18(0.07) &           1.16(0.06) &           1.14(0.04) &           1.38(0.41) &  \textbf{1.13}(0.02) &  1.16(0.00) \\
M4 Hourly                  &           0.86(0.12) &  0.95(0.11) &  \textbf{0.78}(0.07) &           0.85(0.07) &           0.85(0.06) &           1.66(0.00) &  2.66(0.00) \\
M4 Yearly                  &  \textbf{3.05}(0.03) &  3.08(0.04) &           3.05(0.01) &           3.09(0.04) &           3.10(0.02) &           3.38(0.00) &  3.44(0.00) \\
Tourism Quarterly          &           1.61(0.03) &  1.57(0.05) &           1.59(0.05) &           1.59(0.02) &           1.55(0.03) &  \textbf{1.50}(0.01) &  1.83(0.00) \\
Tourism Monthly            &  \textbf{1.42}(0.03) &  1.44(0.03) &           1.45(0.04) &           1.47(0.02) &           1.42(0.02) &           1.44(0.02) &  1.75(0.00) \\
Dominick                   &           0.51(0.04) &  0.49(0.00) &  \textbf{0.49}(0.01) &           0.49(0.01) &           0.49(0.01) &           0.51(0.00) &  0.72(0.00) \\
Kdd Cup                    &           1.20(0.02) &  1.18(0.02) &           1.18(0.03) &           1.18(0.03) &           1.20(0.03) &  \textbf{1.17}(0.01) &  1.39(0.00) \\
Weather                    &           0.63(0.08) &  0.58(0.04) &           0.59(0.02) &           0.59(0.06) &  \textbf{0.57}(0.00) &           0.64(0.01) &  0.69(0.00) \\
NN5 Daily                  &           0.79(0.01) &  0.80(0.01) &           0.81(0.04) &  \textbf{0.78}(0.01) &           0.79(0.01) &           0.86(0.00) &  0.86(0.00) \\
NN5 Weekly                 &           0.76(0.01) &  0.76(0.03) &           0.76(0.01) &           0.77(0.01) &  \textbf{0.76}(0.01) &           0.77(0.01) &  0.87(0.00) \\
Hospital                   &           0.76(0.01) &  0.76(0.00) &           0.76(0.00) &  \textbf{0.75}(0.01) &           0.75(0.01) &           0.76(0.00) &  0.77(0.00) \\
Traffic Weekly             &           1.04(0.07) &  1.10(0.03) &           1.04(0.05) &           1.08(0.09) &           1.03(0.07) &  \textbf{0.99}(0.03) &  1.15(0.00) \\
Electricity Weekly         &           0.78(0.04) &  1.06(0.13) &           0.80(0.04) &  \textbf{0.74}(0.07) &           0.85(0.11) &           0.76(0.01) &  0.79(0.00) \\
Electricity Hourly         &           1.52(0.05) &  1.54(0.00) &           1.58(0.08) &           1.54(0.06) &  \textbf{1.51}(0.05) &           1.60(0.02) &  3.69(0.00) \\
Kaggle Web Traffic Weekly  &           0.56(0.01) &  0.56(0.01) &  \textbf{0.55}(0.00) &           0.57(0.01) &           0.59(0.01) &           0.61(0.02) &  0.62(0.00) \\
Covid Deaths               &           5.11(1.60) &  4.54(0.05) &  \textbf{4.43}(0.13) &           4.58(0.30) &           4.53(0.24) &           5.16(0.04) &  5.72(0.00) \\
Temperature Rain           &           0.76(0.05) &  0.75(0.01) &           0.73(0.02) &           0.73(0.03) &           0.71(0.04) &  \textbf{0.71}(0.03) &  1.23(0.00) \\

\hline
$\varnothing$ Rel. Impr Best &  0.82 &       0.83 &                \textbf{0.81} &                0.81 &                0.82 &                0.86 &         1.0  \\
$\varnothing$ Rel. Impr Oracle &               0.96 &       0.97 &                \textbf{0.95} &                0.95 &                0.96 &                  1.0 &       1.17  \\
\bottomrule
\end{tabular}
}
\caption{We compare variants of \system{} against the single best baseline (TBATS) and a theoretically optimal oracle of choosing the correct baseline for each dataset wrt mean MASE errors on the test sets. We show the mean and standard deviation for each dataset. The best results are highlighted in bold-face. We computed the relative improvement wrt the Oracle Baseline on each dataset and used the geometric average for aggregation over the datasets. }
\label{tab:res_test}

\end{table}

Table~\ref{tab:res_test} shows how different variants of \system{} perform against the two types of baselines across multiple datasets.
Even using the theoretical oracle baseline for comparison, \system{} is able to outperform it on 18 out of 24 datasets. On the other 6 datasets, it achieved nearly the same performance as the baselines. On average, we were able to reduce the MASE by up to $5\%$ against the oracle and by up to $19\%$ against the single best baseline, establishing a new robust state-of-the-art overall. 

Surprisingly, the forecasting-specific budget types did not perform significantly better than the number of epochs (the common budget type in classification). Nevertheless, the optimal choice of budget type varies across datasets, which aligns with our intuition that on a given dataset the correlation between lower and higher fidelities may be stronger for certain budget types than for other types. If we were to construct a theoretically optimal budget-type selector, which utilizes the best-performing budget type for a given dataset, we would reduce the relative error by $2\%$ over the single best (i.e., $\#$~SMPs per Ser.).

\subsection{Hyperparameter Importance}
Although HPO is often considered as a black-box optimization problem~\cite{feurer2019Hyperparameter}, it is important to shed light on the importance of different hyperparameters to provide insights into the design choice of DL models and to indicate how to design the next generation of AutoDL systems. 

Here we evaluate the importance of the hyperparameters with a global analysis based on fANOVA~\cite{Hutter2014fANOVA}, which measures the importance of hyperparameters by the variance caused by changing one single hyperparameter while marginalizing over the effect of all other hyperparameters. Results on individual datasets can be found in appendix. 
\begin{figure}
    \centering
    \includegraphics[width=0.95\textwidth]{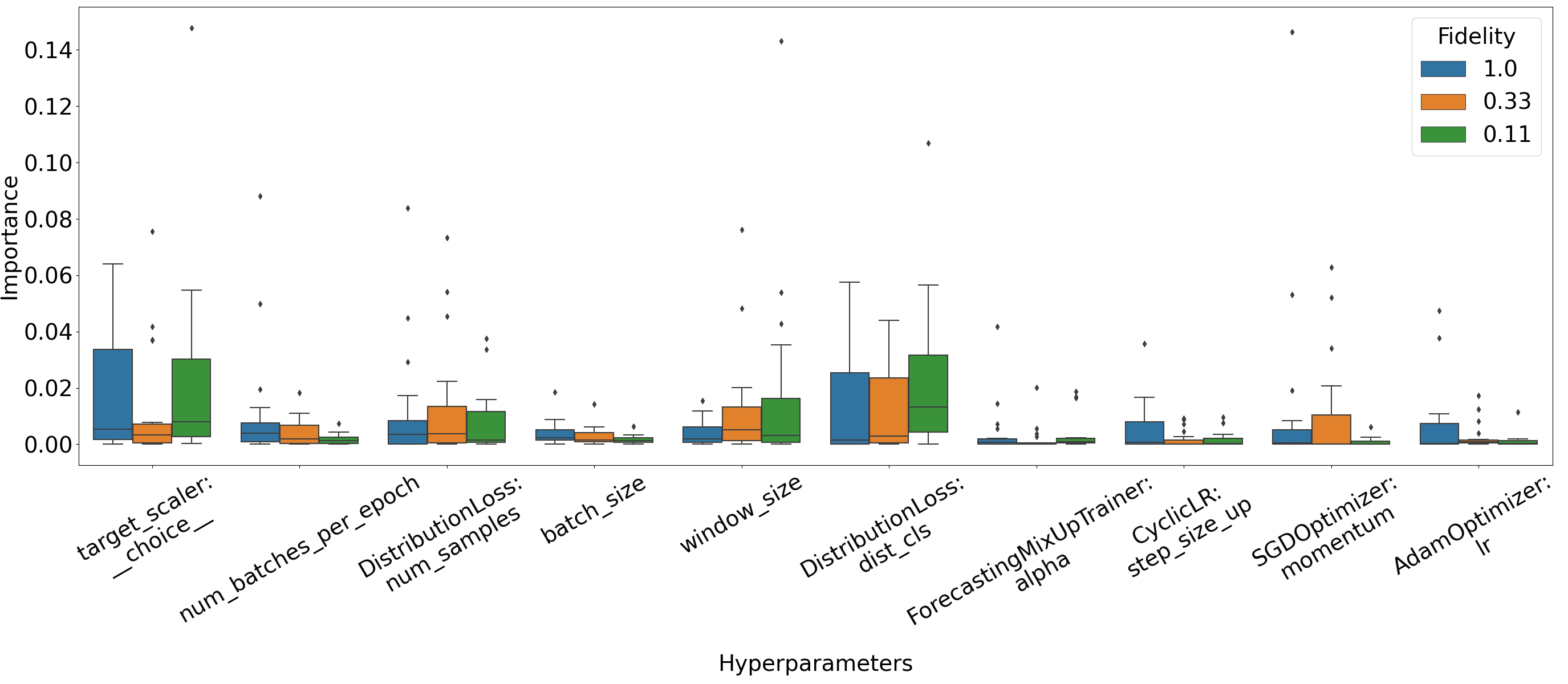}
    \caption{Hyperparameter importance with fANOVA \label{fig:hpimportance} across all datasets of Table~\ref{tab:res_test}}

\end{figure}

For each of the 10 most important hyperparameters in our configuration space (of more than 200 dimensions), Figure~\ref{fig:hpimportance} shows a box plot of the importance across our datasets. 
The most important hyperparameters are closely associated with the training procedure: 3 of them control the optimizer of the neural network and its learning rate. Additionally, 4 hyperparameters (\textit{window\_size}, \textit{num\_batches\_per\_epoch}, \textit{batch\_size}, \textit{target\_scaler}) contribute to the sampler and data preprocessing, showing the importance of the data fed to the network. Finally, the fact that two hyperparameters controlling the data distribution are amongst the most important ones indicates that identifying the correct potential data distribution might be beneficial to the performance of the model.

\subsection{Ablation Study}
\begin{wrapfigure}{hR}{0.57\textwidth}
    \centering
    \includegraphics[width=0.57\textwidth]{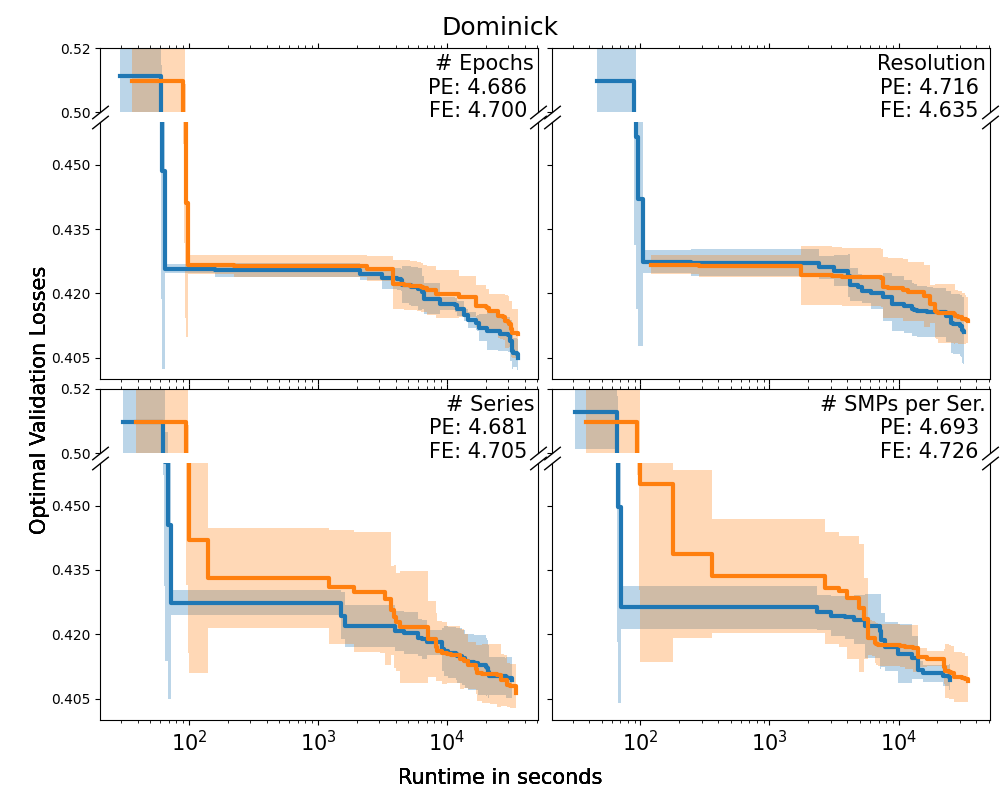}
    \includegraphics[width=0.57\textwidth]{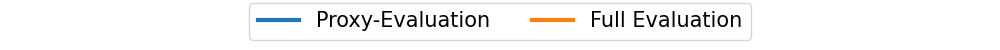}
    \caption{Validation losses over time with different multi-fidelity approaches. We compute the area under the curve (AUC) of our approach (PE) and naive multi-fidelity optimizer (FE) and list them in the figures.
    \label{fig:partial_eval}}
\end{wrapfigure}
In Section~\ref{sec:prox-eval}, we propose to partially evaluate the validation set on larger datasets to further accelerate the optimization process. To study the efficiency gain of this approach, we compare evaluation on the full validation set vs the proxy-evaluation on parts of the validation set. We ran this ablation study on the largest dataset, namely  "Dominick" ($115\,704$ series).

Figure \ref{fig:partial_eval} shows the results. It takes much less time for our optimizer (blue) to finish the first configuration evaluations on the lowest fidelity, improving efficiency early on and showing the need of efficient validation and not only training. We note that the final performance does not change substantially between the different methods. Overall, \system{} achieves the best any-time performance. We note that \system{} has not converged after 10h and will most likely achieve even better performance if provided with more compute resources. The results on the other datasets show a similar trend and can be found in the appendix.

\section{Conclusion and Future Work}

In this work, we introduced \system{}, an AutoDL framework for the joint optimization of architecture and hyperparameters of DL models for time series forecasting tasks. To this end, we propose a new flexible configuration space encompassing several state-of-the-art forecasting DL models by identifying key concepts in different model classes and combining them into a single framework.

Given the flexibility of our configuration space, new developers can easily adapt their architectures to our framework under the assumption that they can be formulated as an encoder-decoder-head architecture. Despite recent advances and competitive results, DL methods have until now not been considered the undisputed best approach in time series forecasting tasks: Traditional machine learning approaches and statistical methods have remained quite competitive~\cite{godahewa2021monash,MAKRIDAKIS2018802M4Anal}. By conducting a large benchmark, we demonstrated, that our proposed \system{} framework is able to outperform current state-of-the-art methods on a variety of forecasting datasets from different domains and even improves over a theoretically optimal oracle comprised of the best possible baseline model for each dataset.

While we were able to show superior performance over existing methods, our results suggest, that a combination of DL approaches with traditional machine learning and statistical methods could further improve performance. The optimal setup for such a framework and how to best utilize these model classes side by side poses an interesting direction for further research. Our framework makes use of BO and utilizes multi-fidelity optimization in order to alleviate the costs incurred by the expensive training of DL models. Our experiments empirically demonstrate, that the choice of budget type can have an influence on the quality of the optimization and ultimately performance.

To the best of our knowledge there is currently no research concerning the choice of fidelity when utilizing multi-fidelity optimization for architecture search and HPO of DL models; not only for time series forecasting, but other tasks as well. This provides a great opportunity for future research and could further improve current state-of-the-art methods already utilizing multi-fidelity optimization. Additionally, we used our extensive experiments to examine the importance of hyperparameters in our configuration space and were able to identify some of the critical choices for the configuration of DL architectures for time series forecasting. Finally, in contrast to previous AutoML systems, to the best of our knowledge, time series forecasting is the first task, where not only efficient training is important but also efficient validation. Although we showed empirical evidence for the problem and took a first step in the direction of efficient validation, it remains an open challenge for future work. \system{} can automatically optimize the hyperparameter configuration for a given task and can be viewed as a benchmark tool that isolates the influence of hyperparameter configurations of the model. This makes our framework an asset to the research community as it enables researchers to conveniently compare their methods to existing DL models.

\section*{Acknowledgements}

Difan Deng and Marius Lindauer acknowledge financial support by the Federal Ministry for Economic Affairs and Energy of Germany in the project CoyPu under Grant No. 01MK21007L.
Bernd Bischl acknowledges funding by the German Federal Ministry of Education and Research (BMBF) under Grant No. 01IS18036A. 
Florian Karl acknowledges support by the Bavarian Ministry of Economic Affairs, Regional Development and Energy through the Center for Analytics – Data – Applications (ADACenter) within the framework of BAYERN DIGITAL II (20-3410-2-9-8).
Frank Hutter acknowledges support by European Research Council (ERC) Consolidator Grant ``Deep Learning 2.0'' (grant no.\ 101045765). \includegraphics[width=0.3\textwidth]{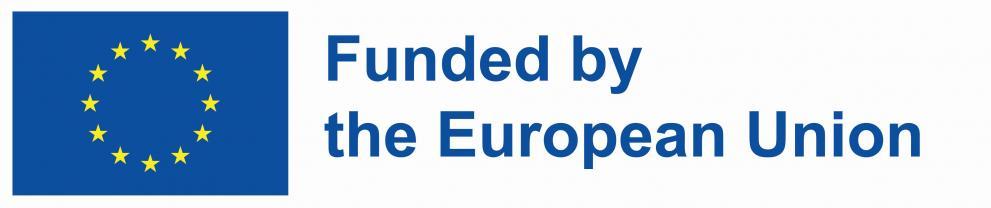}

\end{document}


\appendix
\section{Other Hyperparameters in our Configuration Space}
Besides the choice of the neural architectures, the hyperparameters applied to train a neural network also play a crucial role in the performance of the pipeline. Most of our hyperparameter search space is inherited from Auto-PyTorch for classification ~\cite{zimmertpami21aAPT}\footnote{\url{https://github.com/automl/Auto-PyTorch}}. Here we give a brief overview of the additional forecasting-customized hyperparameters.

Our network is expected to generate one of the following outputs: distribution, quantile or scalar. Network with distribution output is trained with log-probability loss while it can freely select the sort of output distribution (here we implement gaussian and studentT distributions). Network with quantile output is asked to generate a set of output quantiles. Here we only ask the model to forecast the upper bound, median value and lower bound of the target values while the quantiles of upper and lower bound are set as hyperparemeters. Last but not least, networks with scalar output only generate a single value for each time step. Nevertheless, networks with scalar output can be trained with various loss functions, i.e. l1 loss, l2 loss, or mean absolute scaled error (MASE)~\cite{HYNDMAN2006679MASE},. etc. During inference time, we convert the distribution in the following ways: either we take the mean of the distribution as its scalar output, or we sample a certain amount of points from the distribution and take the mean or median values of the samples. All these strategies are considered as hyperparameters that will be optimized by our optimizer. Networks with quantile and scalar output simply forecast with their median value and scalar value respectively.

We implement a sliding window approach to generate the inputs for all the models. The size of the sliding window is heavily dependent on the task at hand, thus we consider the sliding window for the target tasks as a multiple of one $\basewindowsize$. Following ~\cite{HEWAMALAGE2021RNN}, we set the $\basewindowsize$ to be the seasonality period $\period$ (if available) that is no smaller than the forecasting horizon $\horizon$ of the task; if $\horizon$ is greater than all the possible $\period$, we simply take the largest $\period$. As a hyperparameter, the window size ranges from $\basewindowsize$ to $3 \times \basewindowsize$. Additionally, the longest sequence that a CNN can handle is restricted by its receptive field: for TCN models, we simply take their maximal receptive field as the size of the sliding window.

The sliding window approach results in a large amount of overlap between different samples. To avoid overfitting and reduce training time, similar to other frameworks~\cite{gluonts_jmlr}, we restrict the number of  batches at each epoch: the number of training sample instances at each epoch then becomes a fixed value: $batch\_size \times num_{batches}$. We generate the training instances in the following two ways: either each series in the training set is expected to have the same amount of samples or we sample each time step across all the series uniformly. As Auto-PyTorch has already implemented $batch\_size$ as one of its hyperparameters, we simply add the number of batches per epoch and sample strategy as an additional set of hyperparameters.

Neural Networks work best if their input value is bounded. However, unlike tabular datasets where all the data is sampled from the same distribution, the scales of each series in the same dataset can be diversely distributed. Additionally, even the data inside each individual series might not be stationary, i.e., the distribution of the test set might no longer stay in the range of the training/validation sequences. Thus, similar to~\cite{gluonts_jmlr}, we only normalize the data inside each minibatch such that the input of the network is kept in a reasonable range. Similar to other AutoML tools~\cite{FeurerNIPS15ASKL}, data can be scaled in different ways whereas the scaling method is considered as a hyperparameter.  

\section{Hyperparameter Importance for each Dataset}
\begin{figure}
    \begin{subfigure}[b]{0.49\textwidth}
            \centering
            \includegraphics[width=\textwidth]{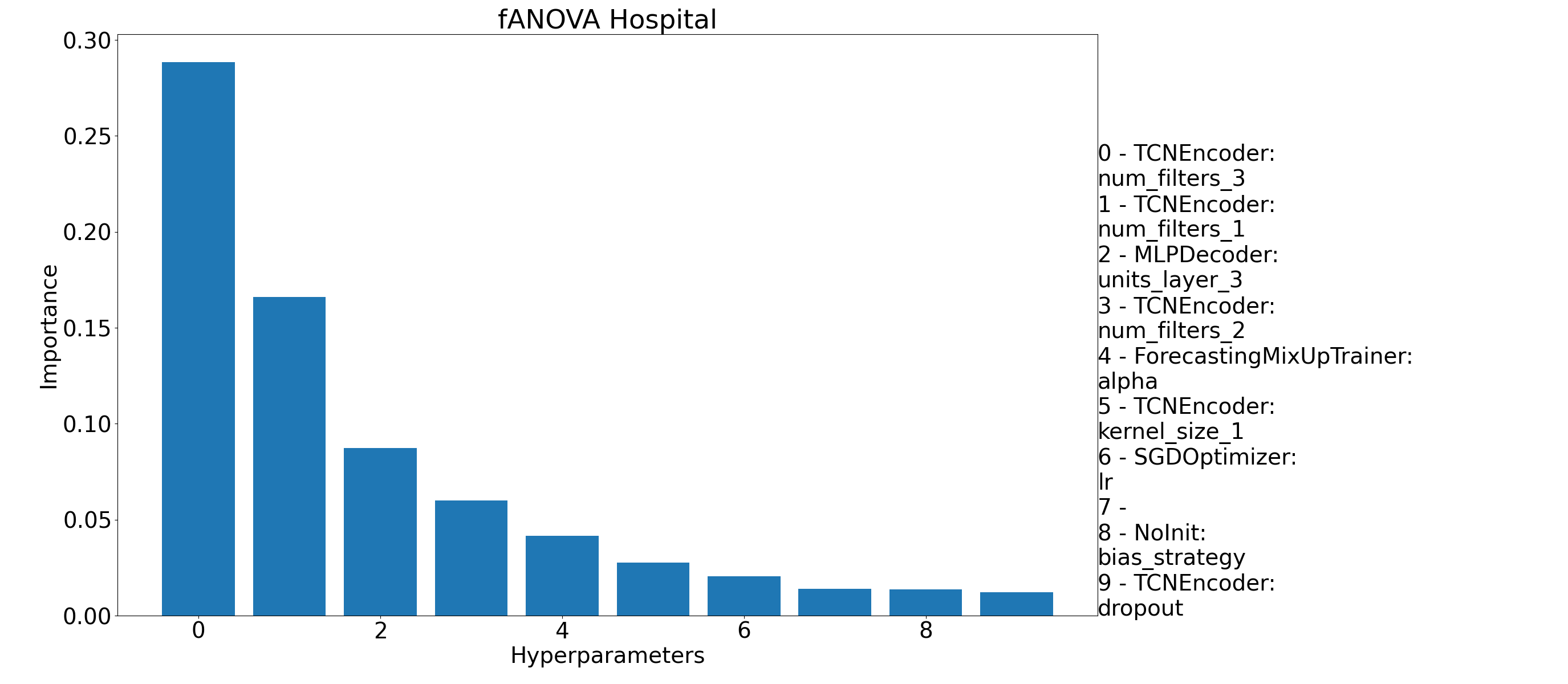}
        \end{subfigure}
        \hfill
        \begin{subfigure}[b]{0.49\textwidth}  
            \centering 
            \includegraphics[width=\textwidth]{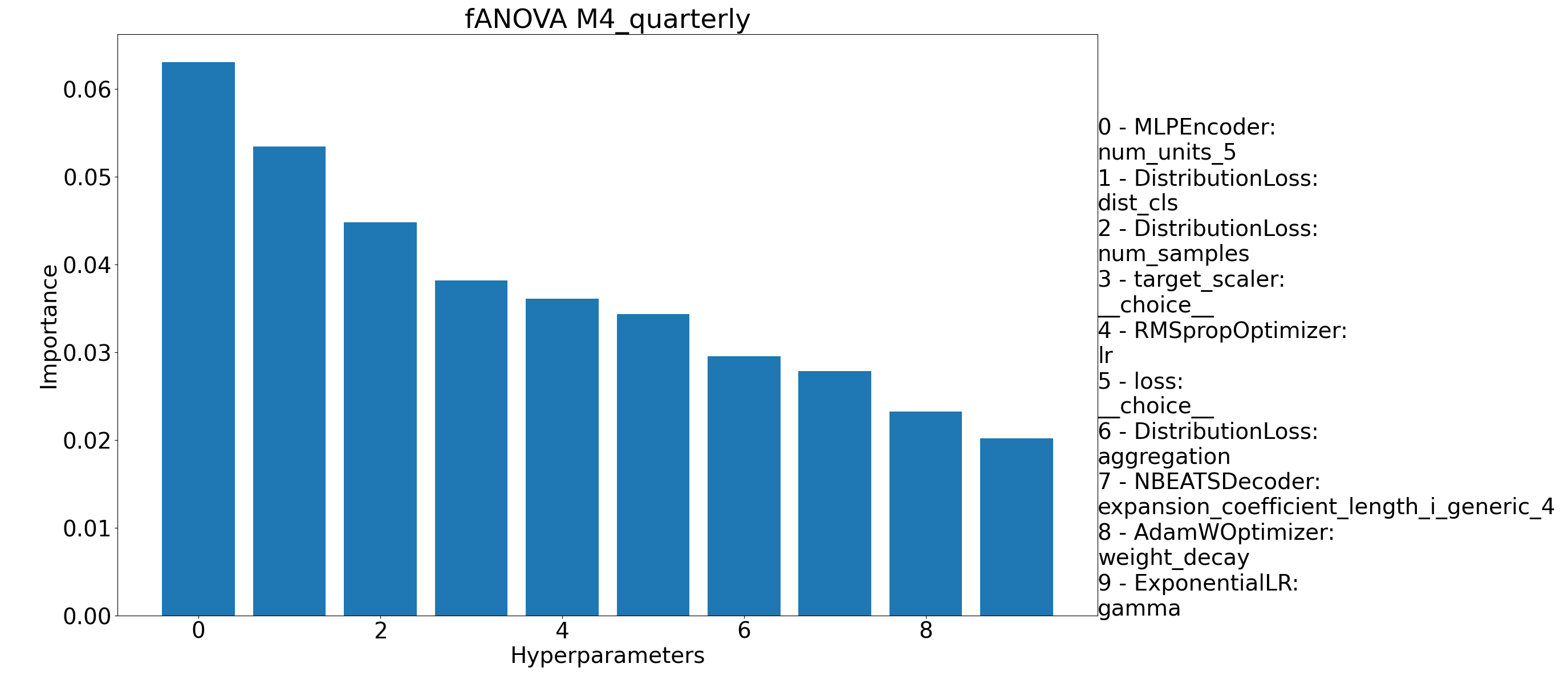}
        \end{subfigure}
        \vskip\baselineskip
        \begin{subfigure}[b]{0.49\textwidth}   
            \centering 
            \includegraphics[width=\textwidth]{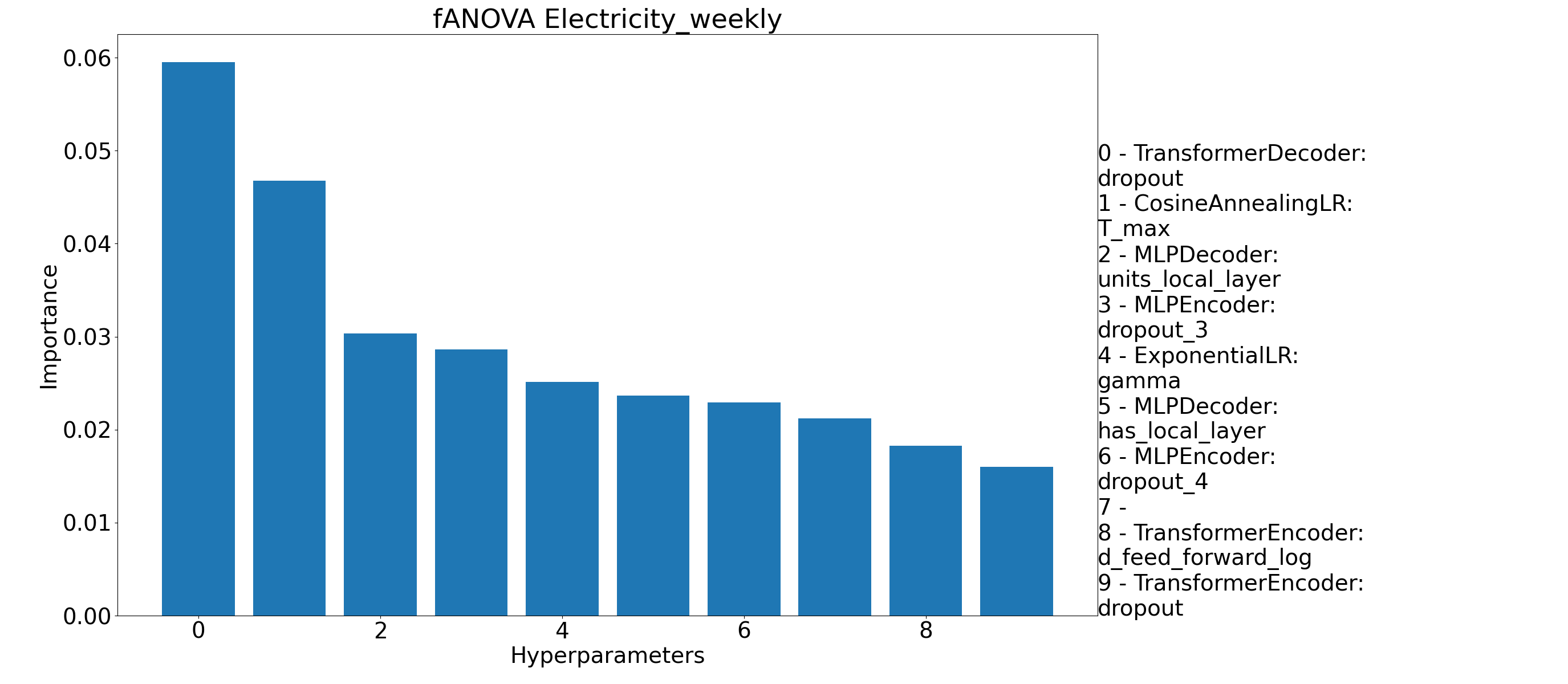}
        \end{subfigure}
        \hfill
        \begin{subfigure}[b]{0.49\textwidth}   
            \centering 
            \includegraphics[width=\textwidth]{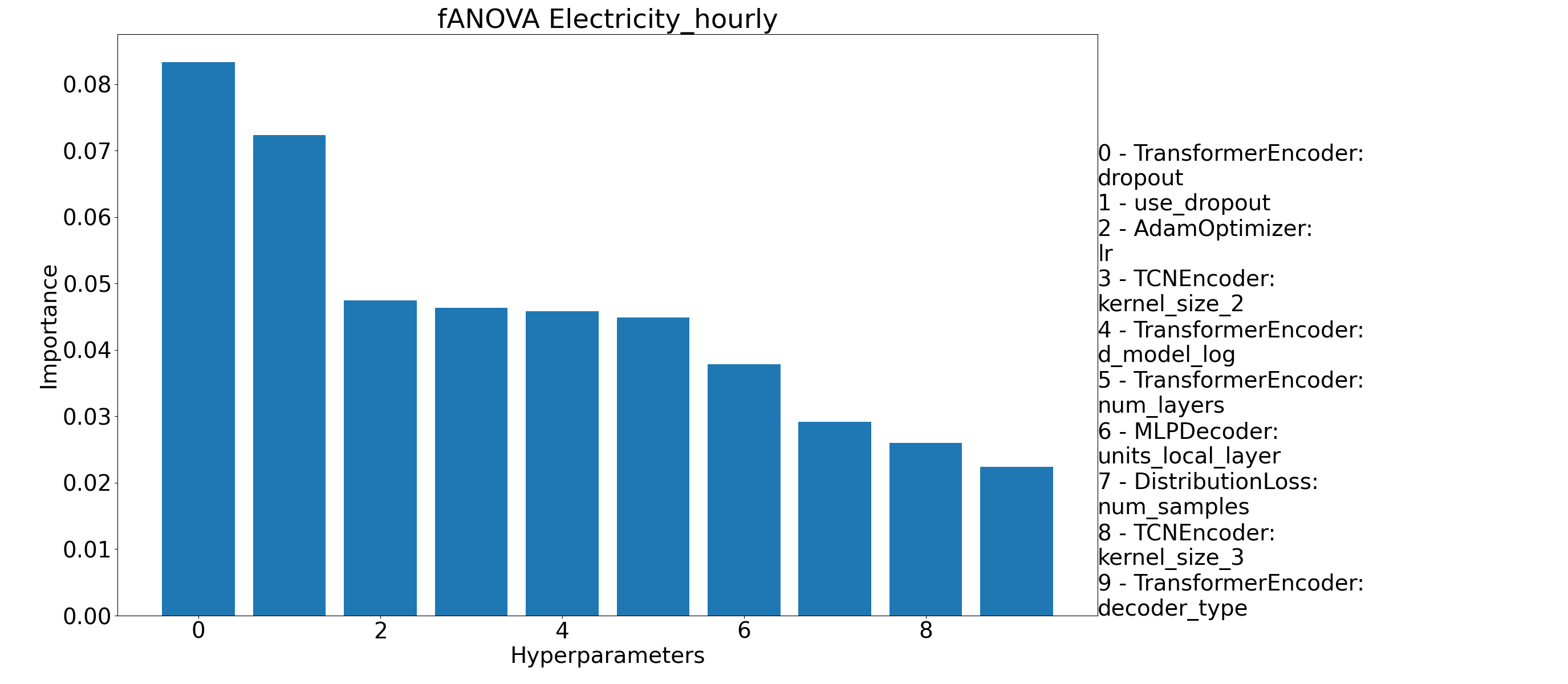} 
        \end{subfigure}
    \caption{Hyperparameter imortance plots based on fANOVA results for four datasets: "Hospital", "M4 quarterly", "Electricity weekly" and "Electricity Hourly".}
    \label{fig:hp_importance_appendix}
\end{figure}

In section 4.2, we compute the importance of all hyperparameters over all the datasets, showing that no single architecture dominates the optimization process. Here we will study the hyperparmeter importance with respect to each individual dataset and evaluate the importance of each hyperparameter. A diverse selection of four datasets is presented in Figure \ref{fig:hp_importance_appendix}. Here we show the hyperparaemter importance on the highest budget (1.0).

It can be seen that architecture-related hyperparameters are among the most important hyperparameters for individual tasks. While different tasks assign different importance values to different architectures. 
To shed a bit of light on the impact of data distribution on hyperparameter importance, we compare "Electricity Weekly" and "Electricity Hourly" side-by-side.
Even comparing these two datasets with similar features from the same domain, differences in hyperparameter importance and preferred architectures can be observed. Both tasks consider the hyperparameters from \textit{Transformer} as the most important hyperparameters. However, "Electricity Weekly" prefer \textit{MLP} as its second important architectures while "Electricity Hourly'' select the hyperparameters from \textit{TCN}, showing that even if the data is sampled from the same distribution, the sample frequency might influence the choice of the optimal architecture.

\section{Further Result on Ablation Study}

\begin{figure}[h!]
    \centering
    \includegraphics[width=0.65\textwidth]{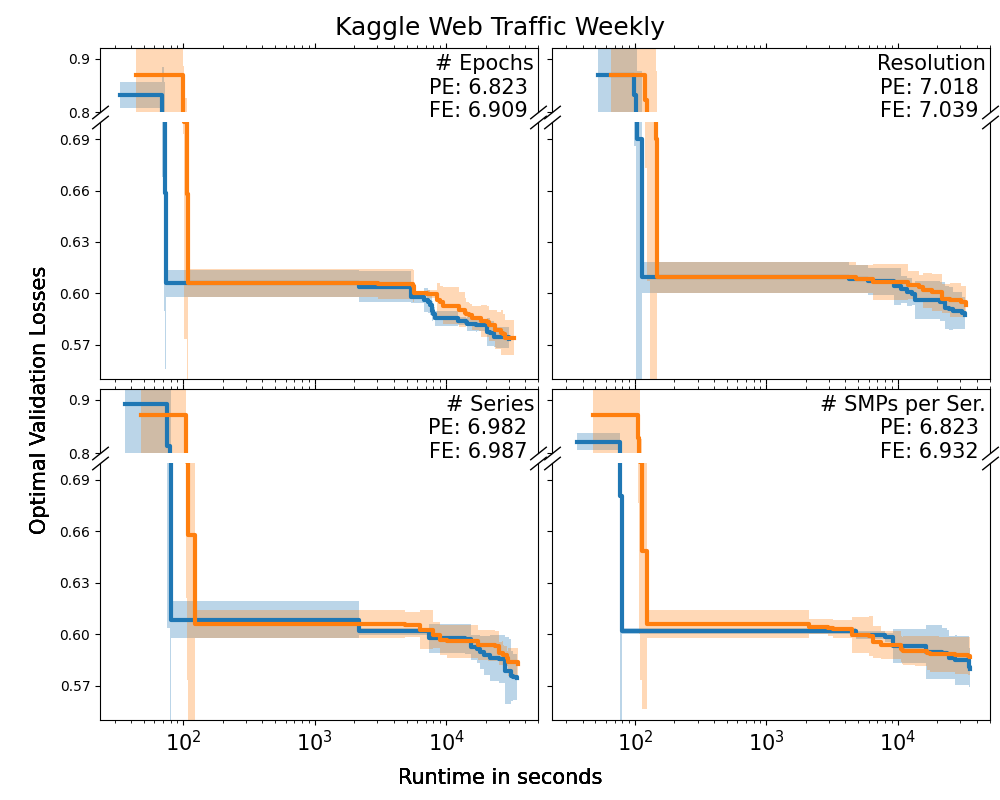}
    \includegraphics[width=0.65\textwidth]{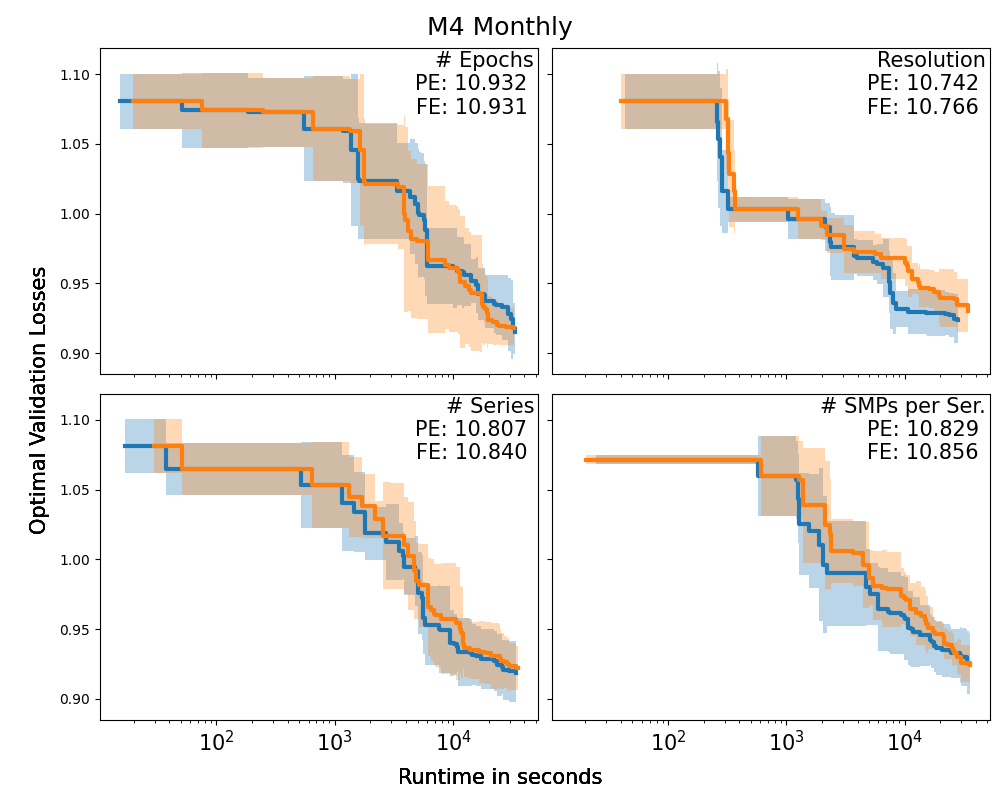}
    \includegraphics[width=0.65\textwidth]{images/ablation/partial_val/ablation_legend.png}
    \caption{Validation losses over time with different multi-fidelity approaches. We compute the area under curves (AUC) of our approach (PE) and naive multi-fidelity optimizer (FE) and attach them in the figure \label{fig:appenix_partial_eval} }
\end{figure}

In Section 4.3, we show that our proxy-evaluation approach helps to achieve a better any-time performance on the "Dominick" ($115\,704$ series) dataset. We show the result on "Kaggle Web Traffic Weekly" ($145\,063$ series) dataset and "M4 Monthly" ($48\,000$) dataset in Figure~\ref{fig:appenix_partial_eval}.
